\documentclass[cameraready]{Interspeech}

\title{From Awareness to Adherence: Bridging the Context Gap in Spoken Dialogue Systems via Context-Aware Decoding}

\author[affiliation={1}, orcid=0009-0007-3419-1895]{Che Hyun}{Lee}
\author[affiliation={3}, orcid=0009-0001-1402-2425, correspondingauthor]{Heeseung}{Kim}
\author[affiliation={1,2}, orcid=0000-0002-2367-197X, correspondingauthor]{Sungroh}{Yoon}

\address{
    $^1$ ECE and $^2$ IPAI, Seoul National University, Seoul 08826, Korea \\
    $^3$ Department of AI, University of Seoul, Seoul 02504, Korea
}

\email{saga1214@snu.ac.kr, gmltmd789@uos.ac.kr, sryoon@snu.ac.kr}


\usepackage{comment}

\usepackage{kotex}
\usepackage{multirow}
\usepackage{multicol}
\usepackage{graphicx}

\begin{document}

\maketitle

\begin{abstract}
    Despite the success of end-to-end (E2E) spoken dialogue systems, maintaining strict context adherence in multi-round conversations remains a challenge. While prior works attribute these failures to models forgetting dialogue history, we highlight an equally critical but overlooked bottleneck: a gap between latent context awareness and active adherence. Although models internally recognize relevant past utterances, strong parametric priors often overshadow these signals during decoding. To bridge this gap, we propose an audio-adapted Context-Aware Decoding (CAD) approach. By leveraging internal attention mechanisms to isolate key historical rounds, our approach contrasts output distributions with and without this key context during inference, directly amplifying multimodal contextual signals. Evaluations on the Audio MultiChallenge benchmark demonstrate significant improvements in Semantic Memory and Self Coherence subtasks, successfully enforcing strict, context-faithful adherence. \footnote{Our code is available at: \url{https://github.com/saga1214/AudioCAD}}
\end{abstract}

\section{Introduction}

End-to-end (E2E) spoken dialogue systems have demonstrated remarkable proficiency in human-computer interaction, specifically in discerning user intent directly from speech and generating contextually appropriate responses \cite{zhang-etal-2023-speechgpt, Qwen-Audio, tang2024salmonn}. Recently, the scope of these systems has expanded from processing single user queries \cite{fang2025llamaomni, kim2024paralinguisticsaware, zeng2025scaling, zhang-etal-2023-speechgpt} to handling complex multi-round communications \cite{chen2024slamomnitimbrecontrollablevoiceinteraction, mitsui-etal-2024-pslm, park-etal-2024-lets, veluri-etal-2024-beyond, zhang-etal-2025-omniflatten}. Unlike processing a single interaction, multi-round communication requires the model to extend beyond immediate utterance processing; it must maintain a coherent dialogue state, allowing the assistant to engage in sustained conversations that feel natural and responsive to the user's evolving needs \cite{chen2025minmomultimodallargelanguage, kyutai2024moshi, fu2025vita, zeng2024glm4, zhong2024lyra}.

However, achieving robust multi-round communication remains a formidable challenge \cite{henderson-etal-2014-second, NEURIPS2023_7b16688a, liu2025voxtral}. In extended dialogues, a system must not only retain the memory of previous user utterances but also ensure consistency with its own prior response \cite{ cieri-etal-2004-fisher, cheng2025omnichatenhancingspokendialogue, park-etal-2024-lets, tong2025interactiveomniunifiedomnimodalmodel, geng2025osum}. Recent benchmarks designed to evaluate these capabilities reveal that current state-of-the-art E2E models often fall short, exhibiting a tendency to \textit{forget} the dialogue history \cite{DBLP:journals/corr/abs-2512-14865, kim-etal-2025-voice-assistant, jung2025avcd}. Consequently, these models often generate responses that are incoherent with the established context or fail to address the user's specific constraints, leading to a degraded user experience where the assistant appears oblivious to the ongoing conversation flow.

Prior studies often address this context modeling failure by assuming that models entirely forget the context, frequently proposing external retrieval-based solutions to supply missing information \cite{10448210, Min2025, 10447448, feng-etal-2025-enhancing, chen-etal-2025-wavrag, lu2025cleans2s}. However, context modeling failures arise not only from severe forgetting but also from an inability to utilize properly retained context. In this work, we focus on the latter case, highlighting a more nuanced bottleneck: a critical gap between \textit{latent awareness} and \textit{active adherence}. We define latent awareness as the model's internal ability to recognize relevant past information, while active adherence refers to the successful manifestation of that context in the final generated output. Although internal representations suggest that models maintain a degree of latent awareness regarding key historical utterances \cite{jain-wallace-2019-attention, zhang2025sentinelattentionprobingproxy, wiegreffe-pinter-2019-attention, yu2025salmonnomni}, these contextual signals are frequently overshadowed by the model's strong parametric priors during the decoding stage. This proves that a significant portion of context failures is not merely a memory deficit, but a generation flaw, necessitating a decoding-centric intervention.

To bridge this gap between latent awareness and active adherence, we propose an audio-adapted Context-Aware Decoding (CAD) approach \cite{shi-etal-2024-trusting}, which amplifies contextual influence by contrasting output distributions generated with and without specific context. While CAD has been utilized in the text domain to mitigate hallucinations \cite{shi-etal-2024-trusting, wang-etal-2025-adacad, khandelwal-etal-2025-cocoa, huangyw-etal-2025-dynamic, 11363442} or reduce recency bias \cite{malkin-etal-2022-coherence}, and recently extended to omni-modal systems to support weak modalities \cite{jung2025avcd}, existing adaptations often rely on coarse heuristics, such as dropping entire histories or using fixed context windows. In contrast, we leverage the model's internal attention mechanisms as a proxy for its latent awareness to dynamically identify and isolate key historical rounds. By specifically penalizing the generic parametric priors associated with these identified rounds during decoding, our precision-targeted approach compels the system to transition from mere internal recognition to strict, context-faithful adherence. Furthermore, we systematically optimize this process through extensive ablation studies on attention layer selection, cross-modal aggregation, and context scoping to precisely calibrate the extraction of latent awareness.

\begin{figure*}[t]
    \centering
    \includegraphics[width=0.9\textwidth]{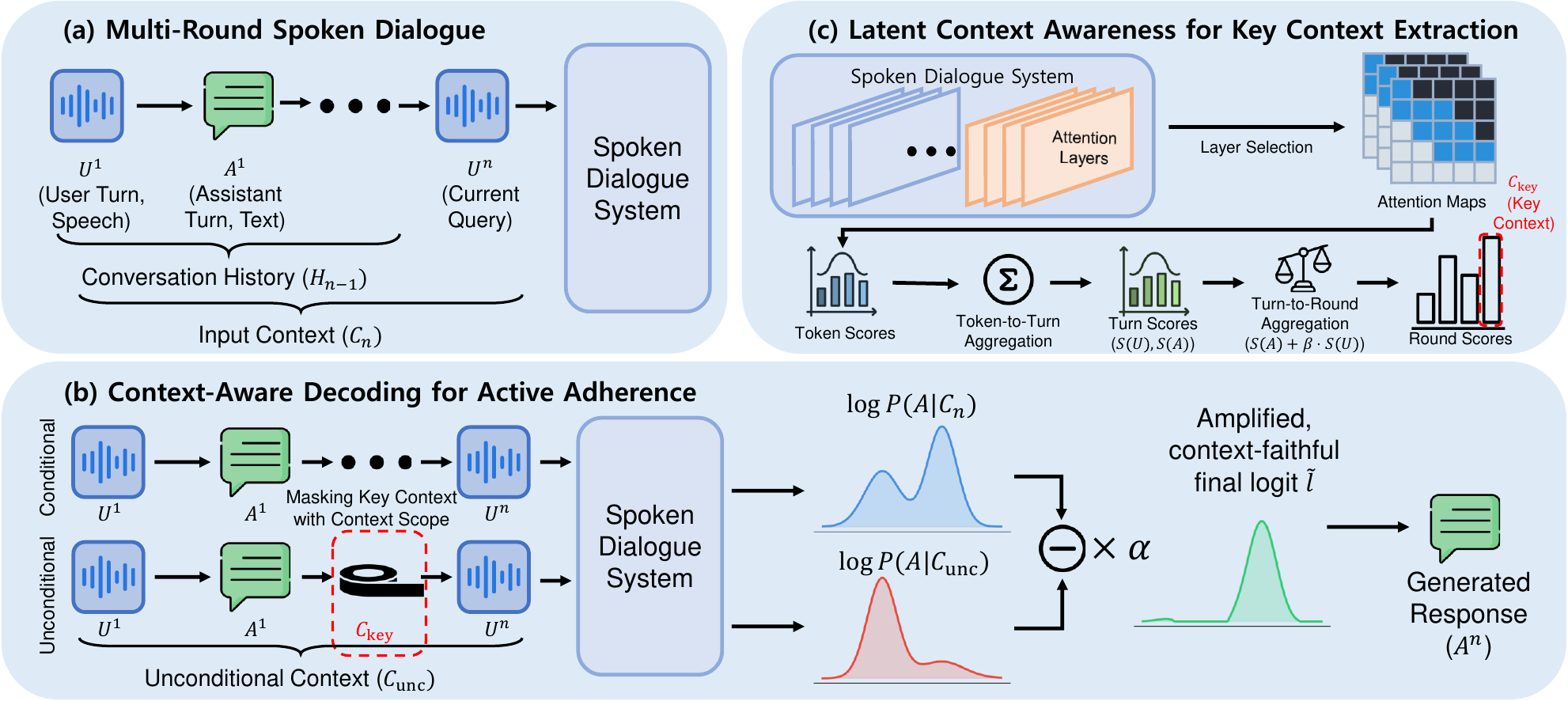}
    \caption{Overall framework of our audio-adapted Context-Aware Decoding (CAD) for spoken dialogue systems. (a) Formalization of multi-round spoken dialogue (Sec.~\ref{sec:multi_round_spoken_dialogue}). (b) CAD mechanism amplifying contextual adherence by penalizing parametric priors during inference (Sec.~\ref{sec:CAD}). (c) Latent awareness pipeline identifying the key context $C_{\mathrm{key}}$ via attention analysis (Sec.~\ref{sec:latent_context_awareness}).}
    \label{fig:method} 
\end{figure*}

In summary, our main contributions are threefold: (1) To the best of our knowledge, this is the first work to explicitly formalize and address the generation failures in multi-round spoken dialogue not merely as a result of forgetting, but as a critical gap between the model's latent awareness and its active adherence to the dialogue context. (2) We propose an audio-adapted Context-Aware Decoding method that amplifies contextual signals directly during inference, effectively mitigating generation flaws without requiring additional training or external retrieval modules. (3) Through extensive evaluations on the Audio MultiChallenge benchmark, we demonstrate that our approach significantly improves strict context adherence, yielding gains in both Semantic Memory and Self Coherence subtasks and thereby effectively reducing hallucinations in extended spoken dialogues.

\section{Methods}

\subsection{Problem Statement: Multi-Round Spoken Dialogue}
\label{sec:multi_round_spoken_dialogue}

End-to-end spoken dialogue systems are fundamentally designed to generate appropriate responses conditioned on user utterances. Recently, the capability of these models has expanded beyond single-round interactions to support sustained, multi-round dialogues. To formally define this process, we first distinguish between a \textit{turn}, which refers to a single utterance from either the user or the assistant, and a \textit{round}, which consists of a paired sequence of a user turn followed by an assistant turn.

Consider a single-round interaction where a user turn is composed of $i$ tokens, denoted as $U = [u_1, u_2, \dots, u_i]$, and the corresponding assistant response consists of $k$ tokens, $A = [a_1, a_2, \dots, a_k]$. An autoregressive spoken dialogue system $\phi$ generates the response by modeling the probability distribution, predicting one token at a time:

\begin{equation}
    P_{\phi}(A|U) = \prod_{j=1}^{k} P_{\phi}(a_j|U, a_{<j}).
\end{equation}

We can extend this formulation to a multi-round scenario, depicted in Fig.~\ref{fig:method}(a). Let the model maintain a conversation history consisting of the previous $n-1$ completed rounds, denoted as $H_{n-1} = [U^1, A^1, \dots, U^{n-1}, A^{n-1}]$. For the $n$-th round, the user provides a new utterance $U^n$. The established input context $C_n$ provided to the model includes the entire past history appended with the current query: $C_n = [H_{n-1}, U^n]$. The assistant must generate the subsequent response $A^n$ conditioned on this extended history:

\begin{equation}
    P_{\phi}(A^n|C_n) = \prod_{j=1}^{|A^n|} P_{\phi}(a^n_j | C_n, a^n_{<j}).
\end{equation}

In general, the tokens of $U$ and $A$ may be either text subwords or speech codecs \cite{kyutai2024moshi}. We focus on a Speech-to-Text (S2T) setting: the user utterance $U$ consists of speech tokens, and the response $A$ is generated as text. Our decoding strategy is nonetheless modality-agnostic and extends directly to Speech-to-Speech (S2S) generation in voice assistants.

The core objective in this setting is to ensure that the generated response $A^n$ is not only contextually appropriate for the immediate query $U^n$ but also strictly faithful and coherent with the past conversation history $H_{n-1}$. Addressing the model's failure to actively adhere to $H_{n-1}$ during the generation of $A^n$ is the main focus of our proposed methodology detailed below.

\subsection{Context-Aware Decoding for Active Adherence}

\label{sec:CAD}

Pre-trained language models frequently suffer from hallucinations in multi-round conversations when explicit dialogue history conflicts with their pre-existing parametric priors. To mitigate this in the text domain, Context-Aware Decoding (CAD) \cite{shi-etal-2024-trusting} was proposed to amplify contextual influence. Standard CAD contrasts the output logits of a model conditioned on the full context against an unconditional setup, which is typically computed by naively dropping the entire history $H_{n-1}$.

We adapt and advance this methodology for the multimodal architecture of spoken dialogue systems, as illustrated in Fig.~\ref{fig:method}(b). Instead of treating the entire history as uniformly relevant, we define a specific key context $C_{\mathrm{key}} \subset H_{n-1}$ capturing the most informative cross-modal signals from speech and text. The detailed pipeline for quantifying latent awareness and extracting $C_{\mathrm{key}}$ is introduced in Sec.~\ref{sec:latent_context_awareness}.

Rather than discarding the full history, we construct our unconditional formulation by masking out only this isolated key context. We define the unconditional context as $C_{\mathrm{unc}} = C_n \setminus C_{\mathrm{key}}$, i.e., the full input with the key context removed. The prediction under $C_{\mathrm{unc}}$ thus reflects the model's parametric prior when it cannot see the key context. Building upon the formalization in Sec.~\ref{sec:multi_round_spoken_dialogue}, we obtain the modified logit $\tilde{l}_j$ for the $j$-th token $a^n_j$ by applying a penalty weight $\alpha$ to this prior. Intuitively, Eq.~\ref{eq:cad} amplifies how much the key context shifts the prediction, pushing the output toward context-favored tokens and away from the prior. We systematically explore $\alpha \in [1.0, 3.0]$ with an increment of $0.5$:

\begin{equation}
\label{eq:cad}
\begin{split}
    \tilde{l}_j &= \log P_{\phi}(a^n_j | C_{\mathrm{unc}}, a^n_{<j}) \\
    &\quad + \alpha \Big[ \log P_{\phi}(a^n_j | C_n, a^n_{<j}) - \log P_{\phi}(a^n_j | C_{\mathrm{unc}}, a^n_{<j}) \Big]
\end{split}
\end{equation} 

\subsection{Latent Context Awareness for Key Context Extraction}
\label{sec:latent_context_awareness}

To construct the key context $C_{\mathrm{key}}$ used in our Context-Aware Decoding formulation, we leverage the \textit{latent context awareness} inherently present within spoken dialogue systems. Although models may fail to manifest historical context in their final output, their internal attention mechanisms often accurately recognize relevant past utterances. We therefore analyze the attention that the current query $U^n$ assigns to each history token: this attention weight serves as the \textit{token score}, indicating how strongly the model attends to that token when forming its response. As illustrated in Fig.~\ref{fig:method}(c), our pipeline then proceeds in stages: we aggregate token scores into turn scores, combine turns into round scores, and finally select the top-scoring rounds as $C_{\mathrm{key}}$. We define this pipeline across four axes:

\textbf{1) Layer Selection:} To determine which layers best capture contextual relevance, we compute token scores from different sets of layers: only the final layer (\texttt{last\_1}), the average of the last four (\texttt{last\_4}), or all layers. This evaluates whether high-level semantics in deeper layers or broader signals across the model better reflect latent awareness.

\textbf{2) Token-to-Turn Aggregation:} To score an entire turn $T \in \{U^m, A^m\}$ rather than a few high-attention tokens, we aggregate the token scores within the turn into a turn-level score $S_{\mathrm{turn}}(T)$, comparing the \texttt{mean}, \texttt{max}, and \texttt{sum}, which respectively capture average, peak, and total attention within the turn.

\textbf{3) Turn-to-Round Aggregation:} A conversation round $R^m$ consists of a user turn (audio codecs) and an assistant turn (text tokens). Because speech is tokenized into far more codecs than the corresponding text, we down-weight the user turn by a turn-to-round ratio $\beta \in \{0.25, 0.5, 1.0\}$ so that a long audio turn does not dominate the round score:
\begin{equation}
    S_{\mathrm{round}}(R^m) = S_{\mathrm{turn}}(A^m) + \beta \cdot S_{\mathrm{turn}}(U^m).
\end{equation}

\textbf{4) Context Scope (Top-$K$ Selection):} Finally, we sort all historical rounds $R^m \in H_{n-1}$ by their $S_{\mathrm{round}}$ scores and take the top $K$ as the key context $C_{\mathrm{key}}$. We investigate $K \in \{1, 2\}$ to analyze the impact of expanding the scope. While a larger $K$ raises the chance of capturing the truly relevant round, it may also introduce spurious signals from irrelevant rounds that degrade our CAD mechanism.

\section{Experiments}

\subsection{Experimental Setup}

\begin{table*}[t]
    \centering
    \setlength{\tabcolsep}{8pt}
    \caption{Performance comparison on the Audio MultiChallenge benchmark. We report the Average Pass Ratio (\%) for Semantic Memory, Self Coherence, and their Average. Applying our proposed Context-Aware Decoding (CAD) consistently improves adherence across all evaluated E2E spoken dialogue systems. ($^*$Note: All scores are uniformly evaluated using \textit{gpt-5-nano} instead of the official \textit{o4-mini} judge, so absolute metrics may differ from the public leaderboard.)}
    \label{table:main_results}
    \begin{tabular}{l | c c c} 
    \toprule
    \multirow{2}{*}{\textbf{Model}} & \multicolumn{3}{c}{\textbf{Average Pass Ratio (\%)$^*$}} \\
    \cmidrule(lr){2-4}
    & \textbf{Semantic Memory} & \textbf{Self Coherence} & \textbf{Average} \\
    \midrule
    MiMo-Audio-7B-Instruct \cite{coreteam2025mimoaudio} & 26.00 & 26.02 & 26.01 \\
    \quad + Ours (CAD)                & \textbf{32.00} & \textbf{36.39} & \textbf{34.11} \\
    \midrule
    Qwen3-Omni-30B-A3B-Instruct \cite{Qwen3-Omni}                       & 22.67 & 29.16 & 25.78 \\
    \quad + Ours (CAD)                & \textbf{39.33} & \textbf{38.80} & \textbf{39.08} \\
    \midrule
    Kimi-Audio-7B-Instruct \cite{kimiteam2025kimiaudiotechnicalreport}           & 13.56 & 19.04 & 16.19 \\
    \quad + Ours (CAD)                & \textbf{23.11} & \textbf{22.65} & \textbf{22.89} \\
    \bottomrule
    \end{tabular}
\end{table*}

\subsubsection{Benchmark and Evaluation}

To evaluate the effectiveness of our CAD approach in multi-round spoken dialogues, we utilize the recently introduced Audio MultiChallenge benchmark \cite{DBLP:journals/corr/abs-2512-14865}. This open-source dataset is specifically designed to assess end-to-end models under natural human interaction patterns, featuring unscripted speech with realistic disfluencies and ambient noise. The conversations span between 3 to 8 continuous rounds, providing a robust testbed for long-horizon state tracking.

Among the evaluation axes, we primarily focus on tasks that require the model to actively adhere to the established context. We select two challenging subsets:
\begin{itemize}
    \item \textbf{Semantic Memory (90 samples):} A subset of the Inference Memory task requiring the assistant to generate responses grounded in the user's past constraints (e.g., explicitly excluding allergens mentioned in earlier rounds).
    \item \textbf{Self Coherence (83 samples):} Evaluates the assistant's ability to maintain consistency with its own prior generations (e.g., consistently recalling and referencing specific items recommended in previous rounds).
\end{itemize}

For evaluation, the benchmark employs a strict, rubric-based methodology. Although we utilize \textit{gpt-5-nano} \cite{singh2025openaigpt5card} as the LLM-as-a-Judge rather than the official benchmark's default evaluator \textit{o4-mini}, we strictly adhere to the rest of the official evaluation pipeline. For each conversation, the judge evaluates the assistant's final utterance against a set of instance-specific rubrics on a binary scale (pass/fail). We report the \textbf{Average Pass Ratio (APR)}, where a generation is considered entirely successful (pass) only if it fully satisfies \textit{all} associated rubrics for given conversation. To ensure statistical reliability, all reported APR metrics are averaged over 5 independent generation runs.

\subsubsection{Baselines}

We evaluate our approach on three state-of-the-art spoken dialogue systems: \textbf{MiMo-Audio-7B-Instruct} \cite{coreteam2025mimoaudio} (with its ``Thinking'' option enabled, MiMo-Audio), \textbf{Qwen3-Omni-30B-A3B-Instruct} \cite{Qwen3-Omni} (Qwen3-Omni), and \textbf{Kimi-Audio-7B-Instruct} \cite{kimiteam2025kimiaudiotechnicalreport} (Kimi-Audio). To isolate the impact of our CAD method, all models utilize their default generation configurations (e.g., $top\_p$, temperature) as specified in their official implementations. We compare standard decoding against our proposed audio-adapted CAD by evaluating the assistant’s response to the final user query in the multi-round dialogue.

\subsection{Results}

Table~\ref{table:main_results} presents the main evaluation results on the Audio MultiChallenge benchmark. For our proposed Context-Aware Decoding, we apply the empirically optimal configuration (\texttt{last\_4}, \texttt{mean}, \texttt{0.5}) derived from Sec.~\ref{sec:abl} across all baseline models.

As shown in the table, integrating our audio-adapted CAD consistently yields substantial performance improvements across all three state-of-the-art spoken dialogue systems. Rather than merely relying on parametric priors, the models equipped with CAD are successfully forced to actively adhere to the established multi-round conversation history. 

Notably, Qwen3-Omni exhibits the most dramatic enhancement, achieving an absolute increase of \textbf{13.30\%} in the APR, with Semantic Memory alone surging from 22.67\% to 39.33\%. Similarly, MiMo-Audio and Kimi-Audio demonstrate significant absolute average gains of \textbf{8.10\%} and \textbf{6.70\%}, respectively. This consistent and robust gain across entirely different models validates that addressing the generation flaw via decoding-centric intervention effectively mitigates hallucination and resolves context failures in multi-turn spoken dialogues.

\subsection{Ablation Studies}
\label{sec:abl}

\begin{figure}[t]
    \centering
    \includegraphics[width=0.95\columnwidth]{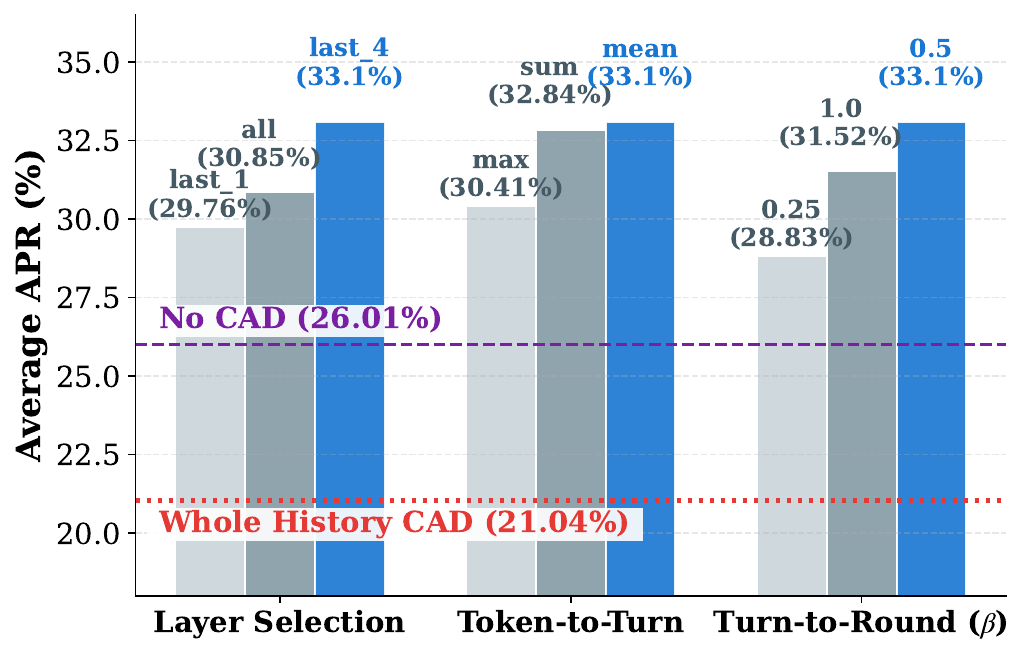}
    \caption{Ablation of CAD hyperparameters: layer selection, token-to-turn aggregation, and turn-to-round aggregation ($\beta$) under a penalty weight of $\alpha=3.0$ and Top-1 context scope. The purple dashed line denotes APR without CAD, and the red dotted line represents Whole History CAD, where the whole history is considered as $C_{\mathrm{key}}$.
    }
    \label{figure:ablation_bar} 
\end{figure}

\begin{figure}[t]
    \centering
    \includegraphics[width=0.95\columnwidth]{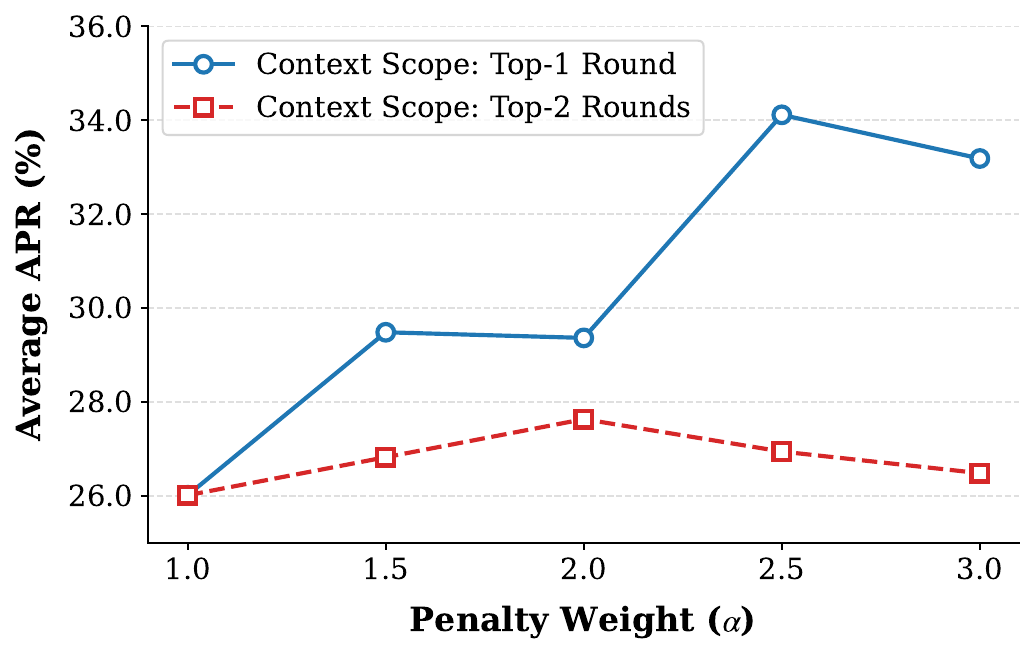}
    \caption{Ablation of CAD hyperparameters: penalty weight ($\alpha$) and context scope ($K$) under the empirically optimal configuration \texttt{last\_4}, \texttt{mean}, and $\beta=0.5$.
    }
    \label{figure:ablation_studies} 
\end{figure}

To validate our design choices and understand the internal dynamics of our method, we conduct comprehensive ablation studies on the CAD hyperparameters. All experiments in this section are performed using MiMo-Audio as a base model, focusing on the key dimensions defined in Sec.~\ref{sec:latent_context_awareness}.

\textbf{Necessity of Latent Awareness:} As shown in Fig.~\ref{figure:ablation_bar}, applying CAD using the whole history $H_{n-1}$ as $C_{\mathrm{key}}$ (\textit{Whole History CAD}) results in a significant performance drop (21.04\%) compared to the \textit{No CAD} baseline (26.01\%). This degradation reveals that naive context selection introduces substantial noise in multi-round spoken dialogues. Our method effectively reverses this trend, achieving 33.10\% by precisely isolating key context.

\textbf{Impact of Layer Selection and Aggregation:} Fig.~\ref{figure:ablation_bar} highlights the performance variations caused by different attention layer selection and aggregation strategies. To clearly isolate these effects, these ablations are evaluated under a fixed penalty weight of $\alpha=3.0$ and a context scope of Top-1 round ($K=1$).

Regarding layer selection, relying solely on the final layer (\texttt{last\_1}) fails to sufficiently capture broader contextual nuances, whereas averaging across all layers introduces excessive noise from low-level features present in the shallow layers. Extracting from the last four layers strikes the optimal balance.

For token-to-turn aggregation, the \texttt{max} overemphasizes isolated outlier keywords within each turn, and \texttt{sum} is biased towards longer turns due to its dependence on the total token count. Consequently, the \texttt{mean} aggregation provides the most stable and representative turn-level score.

Finally, for the turn-to-round ratio ($\beta$), moderately compensating for the inherently longer sequence length of audio tokens ($\beta=0.5$) yields superior results compared to uncompensated ($\beta=1.0$) or overcompensated ($\beta=0.25$) settings. These results confirm that our calibrated setup optimally captures the latent awareness and successfully isolates the key context.

\textbf{Impact of Penalty Weight and Context Scope:} We further investigate the effect of the penalty weight ($\alpha \in \{1.0, 1.5, 2.0, 2.5, 3.0\}$) across different context scopes ($K \in \{1, 2\}$), as illustrated in Fig.~\ref{figure:ablation_studies}. For a context scope of 1 round ($K=1$), increasing the penalty weight steadily improves the Average APR until the performance eventually saturates at higher values, leading us to select $\alpha=2.5$ as the optimum.

Conversely, expanding the context scope to 2 rounds ($K=2$) generally degrades the overall performance. Notably, performance drops sharply as $\alpha$ increases. This indicates that a larger scope likely incorporates irrelevant turns. Aggressively amplifying such irrelevant context via high penalty weights is detrimental, underscoring the need for precise context selection.

\section{Conclusion}

In this work, we introduced an audio-adapted Context-Aware Decoding (CAD) framework to bridge the inherent context gap between latent context awareness and active adherence in multi-round spoken dialogues. Our comprehensive evaluation on the Audio MultiChallenge benchmark demonstrates that our proposed approach consistently yields substantial performance improvements across diverse state-of-the-art spoken dialogue systems. These results validate that decoding-centric interventions serve as a robust, model-agnostic strategy for enhancing the faithfulness of spoken dialogue systems.

\section{Generative AI Use Disclosure}

Generative AI tools were utilized only for language editing and polishing. The final content was produced and reviewed entirely by the authors.

\section{Acknowledgments}

This work was supported by Institute of Information \& Communications Technology Planning \& Evaluation (IITP) grants funded by the Korea government (MSIT) [NO.RS-2021-II211343, Artificial Intelligence Graduate School Program (Seoul National University); No.2022-0-00959, RS-2022-II220959], by National Research Foundation of Korea (NRF) grant [No.2022R1A3B1077720, 2022R1A5A7083908], BK21 FOUR Program of the Education and Research Program for Future ICT Pioneers, Seoul National University in 2026. This was supported by Mobile eXperience(MX) Business, Samsung Electronics Co., Ltd.

\bibliographystyle{IEEEtran}
\bibliography{mybib}

\end{document}